\documentclass[runningheads]{llncs}
\usepackage[T1]{fontenc}
\usepackage{cite}
\usepackage{amsmath,amssymb,amsfonts}

\usepackage{graphicx}
\usepackage{textcomp}
\usepackage{xcolor}
\usepackage{booktabs} 
\usepackage{siunitx}  
\usepackage{multirow} 
\usepackage{algorithm}
\usepackage{algpseudocode}
\usepackage{caption}
\begin{document}
\title{A Target-based Multi-LiDAR Multi-Camera Extrinsic Calibration System}
%
%
\author{Lorenzo Gentilini\inst{1}\and
Pierpaolo Serio\inst{2,3} \and
Valentina Donzella\inst{3}\and
Lorenzo Pollini\inst{2}}
\authorrunning{L. Gentilini et al.}
%
\institute{Research \& Development, Toyota Material Handling Manufacturing, Bologna Italy\\
\email{lorenzo.gentilini@toyota-industries.eu}\\ \and
Department of Information Engineering, University of Pisa, Pisa, Italy
\email{pierpaolo.serio@phd.unipi.it}\\
\email{lorenzo.pollini@unipi.it}\\ \and
School of Engineering and Materials Science, Queen Mary University of London, London, United Kingdom\\
\email{v.donzella@qmul.ac.uk}}

\maketitle              
\begin{abstract}
Extrinsic Calibration represents the cornerstone of autonomous driving. Its accuracy plays a crucial role in the perception pipeline, as any errors can have implications for the safety of the vehicle. Modern sensor systems collect different types of data from the environment, making it harder to align the data. To this end, we propose a target-based extrinsic calibration system tailored for a multi-LiDAR and multi-camera sensor suite. This system enables cross-calibration between LiDARs and cameras with limited prior knowledge using a custom ChArUco board and a tailored nonlinear optimization method. We test the system with real-world data gathered in a warehouse. Results demonstrated the effectiveness of the proposed method, highlighting the feasibility of a unique pipeline tailored for various types of sensors.

\keywords{Navigation, Perception \& SLAM \and Advances in Sensor Technology \and Mechatronic systems}
\end{abstract}
\section{Introduction}
Perception represents the bridge between the environment and the autonomous vehicle's processing pipeline. The information from the vehicle's surroundings has to be collected taking into account the relative positioning of the sensors on the vehicle body. In fact, in the absence of accurate calibration, even minor misalignments can lead to erroneous obstacle localization or scene misinterpretation, thereby significantly compromising the reliability of subsequent decision-making processes. Nowadays, Light Detection and Ranging (LiDAR) and camera systems are attracting more interest due to the effective combination of vision and ranging sensors with high resolution\cite{li2020lidar, hsu2020review}.
\begin{figure}
    \centering
    \includegraphics[width=\columnwidth]{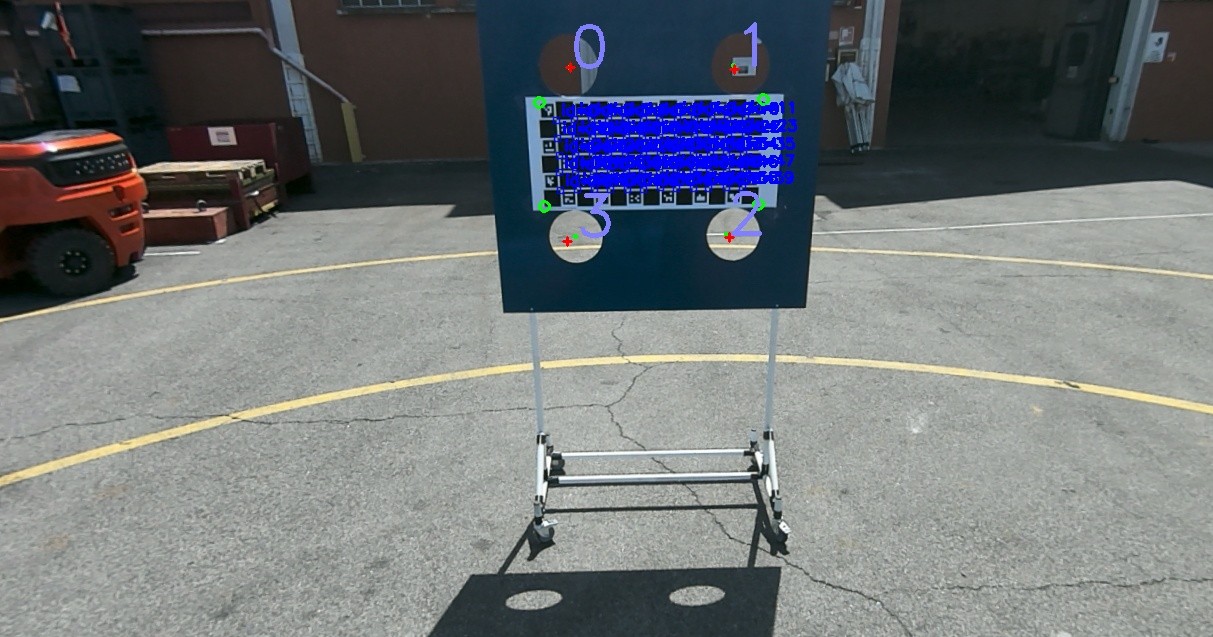}
    \caption{Projection of the board centers found by a LiDAR (red cross) and a camera (green dot). Blue IDs are the ChArUCo marker detections from the camera. }
    \label{fig:lidarcamera_centers_projection}
\end{figure}

This choice aims to take the best of both worlds: the depth estimation accuracy of a LiDAR enriches the substantial amount of semantic information coming from the Camera. To this end, an effective extrinsic calibration would ensure a smooth overlap between the 3D projection of the 2D pixel information and the LiDAR point cloud. However, in this particular task, it should be taken into account the the inherent heterogeneity of the data specific to this task makes harder to apply a universal solution. The challenge is further exacerbated by the variability in sensor's specifications and noise characteristics, all of which affect the robustness of calibration. In this context, we propose a novel system that utilizes a custom anchor board, designed to be detected by both LiDAR and camera. After detection, the calibration can be formulated as an optimization problem that computes the transformation required to align every measurement using the detected board for each sensor combination. The presented system was tested using real-world data collected in a warehouse. In sum, the main contribution of this work lies in the design and implementation of a calibration framework with a focus on scalability, tailored for multi-LiDAR multi-Camera sensor suites that leverages a custom-designed anchor board and formulates an extrinsic calibration as a unique optimization problem for the whole sensor suite that provides the mutual transformation between every sensor, even between sensors with no overlapping Field of View. By addressing the heterogeneity of sensor data and ensuring reliable cross-modal correspondence, the proposed method enhances the accuracy and robustness of multi-sensor integration in real-world scenarios. The paper is structured as follows: Section \ref{section:relworks} introduces the state-of-the-art of extrinsic calibration, Section \ref{section:problemstatement} presents a formal problem formulation, Section \ref{section:ourapproach} describes the proposed approach, and Section \ref{section:experiments} shows the experiment results.

\section{Related Works}\label{section:relworks}
Several paradigms for extrinsic multi-LiDAR multi-Camera calibration have emerged over time. While the research of extrinsic calibration for the same type of sensor can rely on a wide range of solutions, the development of a versatile multi-sensor system remains an active area of study. Integrating sensors with varying modalities, resolutions, and fields of view introduces considerable complexity in achieving precise spatial alignment.
A broad division is between target-based and target-less approaches. 

\subsection{Targetless methods}
Targetless approaches do not require artificial elements in the environment, as they extract relevant information from their surroundings to perform extrinsic calibration. In this context, various types of information can be utilized. Range data can be registered exploiting their global appearance, that is, to find the best transformation that overlaps the two pointclouds\cite{du2025,shahbeigi2024}.
In scenes populated with numerous landmarks, it is possible to extract relevant features from both the visual and the range data. In \cite{ma2021crlf}, the authors propose a novel, fully automatic, and user-friendly method for LiDAR-camera extrinsic calibration in road scenes, leveraging line features from static straight-line-shaped objects. In a similar vein, \cite{lin2025} proposes a novel online target-less LiDAR-camera calibration method that combines geometric edges and semantic features extracted using \cite{kirillov2023segment} (e.g., road markings, traffic signs) with a robust feature selection strategy based on contour matching across multiple frames. Likewise, \cite{yin2023} presents an automatic and target-less calibration method that leverages motion consistency and edge alignment between LiDAR and camera data in natural scenes. Additionally, in \cite{kang2020automatic}, the authors introduce a differentiable calibration method based on a Gaussian Mixture Model that aligns many-to-many correspondences between LiDAR and image edges using an edge-aware cost function.

\begin{figure}[tbp]
    \centering
    \includegraphics[width=0.4 \columnwidth]{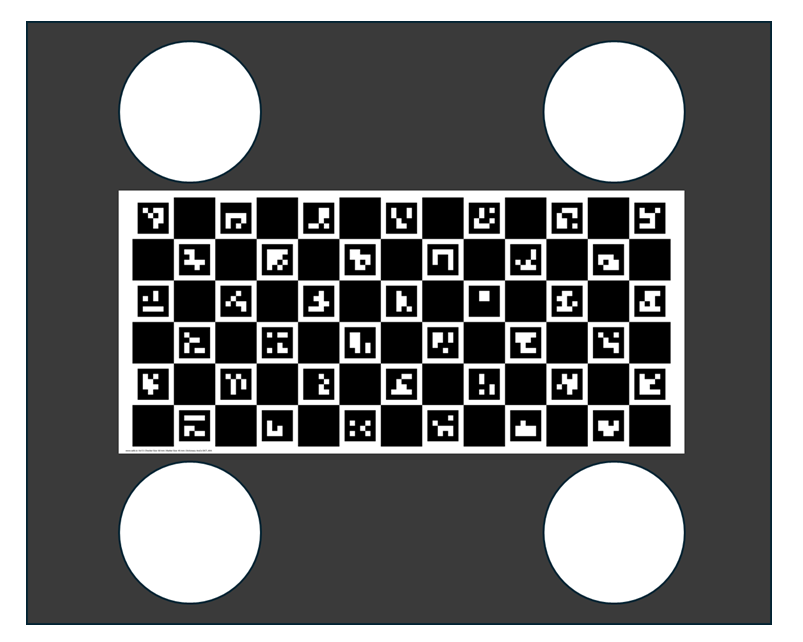}
    \caption{Design of the target board.}
    \label{fig:targetrender}
\end{figure}

\subsection{Target-based methods}
Finding shared, naturally occurring features across all sensor types to calibrate them simultaneously is difficult because each sensor operates on a different principle. Target-based algorithms are investigated to overcome such issues. The golden standard for camera calibration targets is a planar board with distinctive patterns, such as a checkerboard, due to its reliable visibility across diverse environments \cite{an2024survey, yaopeng2021review}. The system presented in \cite{zhou2018automatic}  implements a calibration method for 3D LiDAR and camera systems using line and plane correspondences from a single checkerboard pose.
In \cite{park2014}, the authors leverage a polygonal planar board, estimating 3D-2D point correspondences from board vertices to compute the projection matrix without separate intrinsic and extrinsic steps. To make the board detection more robust to occlusions and partial occlusions, it can be enriched with ChArUco markers\cite{an2018charuco}. In \cite{beltran2022automatic}, the target board includes four ArUco markers to improve the camera detection and four holes to enable the detection from the LiDAR.

\section{Proposed Approach}\label{section:problemstatement}
\subsection{Problem Statement}
The system presented in this work performs extrinsic calibration between multiple LiDAR sensors and cameras. This is achieved by simultaneously exploiting images of a calibration target acquired by the cameras and its geometric information extracted from the LiDAR sensors. In the case of calibration between heterogeneous sensors, the target images are appropriately projected onto the corresponding geometric data.To formalize the problem, we define a sensor suite  $(L_1, \dots,L_n,C_1,\dots,C_m)$, consisting of $n$ LiDAR sensors and $m$ cameras. For each sensor, we associate a reference frame in which data are acquired, denoted as ${L^i}$ for the $i^{th}$ LiDAR and ${C^j}$ for the $j^{th}$ camera. Solving the extrinsic calibration between ${L^i}$ and ${C^j}$ consists in estimating the transformation $T_{L_i}^{Cj}$, which defines the relative pose between the two sensors.

Our calibration system leverages the data acquired from each sensor in the suite by systematically exploiting their respective characteristics across all possible sensor pair combinations. For clarity and consistency in notation, we denote a generic LiDAR sensor by \( L \) and a generic camera by \( C \), both drawn from a heterogeneous sensor suite. The LiDAR sensor \( L \) generates a point cloud comprising \( N \) points, expressed in its local coordinate frame as
\[
^{L}\mathbf{P} = [ \mathbf{x}\; \mathbf{y} \; \mathbf{z} ]^T \in \mathbb{R}^{3 \times N},
\]
where each column corresponds to the 3D coordinates of a single point.

The camera sensor \( C \) captures a 2D image \( I \in \mathbb{R}^{H \times W \times 3} \), where \( H \) and \( W \) represent the image height and width, respectively. Pixel coordinates within this image are denoted as \( [u, v] \in \mathbb{R}^2 \), expressed in the image plane.

The goal of extrinsic calibration is to estimate the rigid-body transformation \( \mathbf{T}_{L}^{C} \in SE(3) \) that maps the LiDAR coordinate frame to the camera frame. This transformation consists of a rotation matrix \( \mathbf{R} \in SO(3) \) and a translation vector \( \mathbf{t} \in \mathbb{R}^3 \), such that:
\[
\mathbf{T}_{L}^{C} = \begin{bmatrix} \mathbf{R} & \mathbf{t} \\ \mathbf{0}^\top & 1 \end{bmatrix}.
\]

The calibration process involves optimizing \( \mathbf{T}_{LC} \) to minimize a cost function based on the reprojection error between the observed 2D locations of anchor board features in the camera image and the projections of their corresponding 3D locations in the LiDAR point cloud. This optimization framework ensures a consistent and accurate alignment across all sensor pairs, thereby enhancing the overall integrity of multi-sensor fusion.

\begin{figure*}[htbp]
    \centering
    \includegraphics[width=\textwidth]{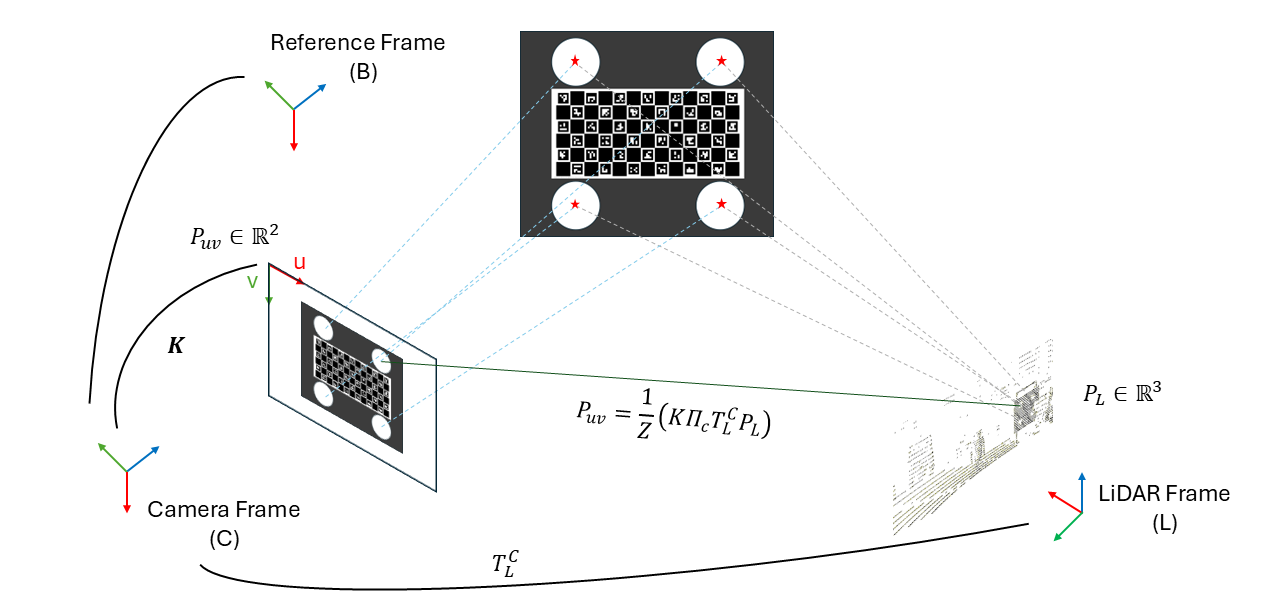}
    \caption{Calibration scheme of a generic Camera-LiDAR system with related frames.}
    \label{fig:calibrationscheme}
\end{figure*}

\subsection{Target}\label{section:ourapproach}
To align the data perceived by both types of sensors, we employ an artificial anchor that serves as the reference object for alignment. To ensure the highest possible accuracy in this operation, we used the calibration target shown in Figure \ref{fig:targetrender}. This target consists of a checkerboard with ArUco markers and circular patterns located at its corners. ChArUco boards combine a classical checkerboard layout with embedded ArUco markers, which are uniquely identifiable binary codes placed within the white squares. This unique encoding enables the individual recognition of each light square, even under partial occlusion or suboptimal lighting conditions. As a result, the detection algorithm can reliably identify and use the visible markers and corners for accurate calibration. The circular patterns at the corners serve as additional markers that can also be detected by the LiDAR. Specifically, the LiDAR perceives these circles as significant discontinuities in depth measurements (i.e., as voids or holes), while for the camera, they can be interpreted as fixed offsets relative to the corners of the ChArUco board.

\subsection{Target detection}

To detect the target, the system employs the ChArUco board detection function from the OpenCV library on the images captured by the cameras. Once the cameras observing the target are identified, the system proceeds to solve the \textit{target pose computation problem}. This problem can be formulated as a Perspective-n-Point (PnP) problem, which involves estimating the 3D pose of the target based on its 2D image projection. Given the intrinsic parameters \([\phi_x, \phi_y]\) and \([c_x, c_y]\), the PnP algorithm computes the rotation and translation components that satisfy the following relation:
\begin{align}
\begin{bmatrix}
u \\ v \\ 1
\end{bmatrix}
=
\frac{1}{Z} \left(
\mathbf{K} \, \mathbf{\Pi}_c
\begin{bmatrix}
r_{11} & r_{12} & r_{13} & t_x \\
r_{21} & r_{22} & r_{23} & t_y \\
r_{31} & r_{32} & r_{33} & t_z
\end{bmatrix}
\begin{bmatrix}
X \\ Y \\ Z \\ 1
\end{bmatrix}
\right)
\end{align}

where

\begin{align*}
\mathbf{K} =
\begin{bmatrix}
\phi_x & 0 & c_x \\
0 & \phi_y & c_y \\
0 & 0 & 1
\end{bmatrix}
\quad \text{and} \quad
\mathbf{\Pi}_c =
\begin{bmatrix}
1 & 0 & 0 & 0 \\
0 & 1 & 0 & 0 \\
0 & 0 & 1 & 0
\end{bmatrix}.
\end{align*}

\begin{algorithm} 
\caption{Cloud Processing and Circle Refinement for LiDARs}
\begin{algorithmic}
\Require Point cloud $\mathcal{C}$, initial transformation $\mathbf{T}_{\text{init}}$, target mask $\mathcal{T} $
\Ensure Refined transformation $\mathbf{T}_{\text{refined}}$, refined circle centers $\mathcal{P}_{\text{circles}}$

\State \textbf{Filter point cloud:}
\State $\mathcal{C}_f \gets \{ \mathbf{p} \in \mathcal{C} \mid z_p \geq h_{\min} \land d_{\min} < \|\mathbf{p}_{xy}\| \leq d_{\max} \}$

\State \textbf{Align calibration target to filtered cloud (GICP):}
\State Apply $\mathbf{T}_{\text{init}}$ to pointcloud target model $\mathcal{T} \rightarrow \mathcal{T}'$ and $\mathcal{C}_f$
\State Register transformed $\mathcal{T}'$ to filtered LiDAR measurement $\mathcal{C}_f$ via GICP
\If{GICP converges and fitness $< \varepsilon$}
    \State $\mathbf{T}_{\text{refined}} \gets \mathbf{T}_{\text{GICP}} \cdot \mathbf{T}_{\text{init}}$
\Else
    \State \Return failure
\EndIf

\State \textbf{Extract matched points via nearest neighbor:}
\State $\mathcal{C}_{\text{match}} \gets \{ \mathbf{p} \in \mathcal{C}_f \mid \exists \mathbf{q} \in \mathcal{T}' \text{ with } \|\mathbf{p} - \mathbf{q}\| < \delta \}$

\State \textbf{Segment calibration plane (RANSAC):}
\State Fit plane $\pi: a x + b y + c z + d = 0$ to $\mathcal{C}_{\text{match}}$
\State Keep inliers $\mathcal{P}_\pi$ such that $\text{dist}(\mathbf{p}, \pi) < \epsilon$

\State \textbf{Normalize plane to horizontal:}
\State Compute angles $\theta_x = \arctan(b/c), \theta_y = -\arctan(a/c)$
\State Compute rotation matrix $\mathbf{R}_{\text{plane}} = \mathbf{R}_x(\theta_x) \cdot \mathbf{R}_y(\theta_y)$
\State Apply $\mathbf{R}_{\text{plane}}$ to $\mathcal{P}_\pi \rightarrow \mathcal{P}_{\text{rot}}$

\State \textbf{Build 2D occupancy grid:}
\State Define grid resolution $r = 200$ cells/m
\State Project $\mathcal{P}_{\text{rot}}$ into grid $\mathcal{O}(i,j)$
\State $\mathcal{O}(i,j) = \text{true} \iff \exists \mathbf{p} \in \mathcal{P}_{\text{rot}}$ within cell

\State \textbf{Find densest target region:}
\State Slide window of size $s \times s$ (target size)
\State Find $(i^*, j^*) = \arg\max \sum_{i,j \in \text{window}} \mathcal{O}(i,j)$

\State \textbf{Estimate and refine circle centers:}
\ForAll{circle offsets $\mathbf{o}_k$}
    \State Initial centers positions $\mathbf{c}_k \gets$ $(i^*, j^*) + \mathbf{o}_k$
    \State Refine centers positions $\mathbf{c}_k$ by minimizing background pixels in circular mask
\EndFor
\State $\mathcal{P}_{\text{circles}} \gets \{ \mathbf{c}_k \}$

\State \textbf{return} $\mathbf{T}_{\text{refined}}, \mathcal{P}_{\text{circles}}$
\end{algorithmic}
\label{pseudocode}
\end{algorithm}

Here, \((X, Y, Z)\) denotes a point on the target in the world coordinate system, and \((u, v)\) is its corresponding image coordinate in the 2D image plane. The matrices \(\mathbf{K}\) and \(\mathbf{\Pi}_c\) represent the camera's intrinsic matrix and the projection operator, respectively. The transformation matrix encodes the rotation and translation parameters that define the target's pose relative to the camera frame. In the target detection with LiDAR sensor exploits holes in the board. First, the system filters out the ground points, which interfere with the following alignment steps. Then it matches the pointcloud with a target mask to get an approximate initial guess using Generalized Iterative Closest Point (GICP) algorithm \cite{gicp2009}. Once that an initial guess is provided, a Random Sample Consensus (RANSAC) algorithm  \cite{fischler1981random} filters out everything but the target board and aligns its surface with the one of the masks.
To get the final center positions, the system leverages a 2D occupancy map-based representation with a 200 points-per-meter resolution. In this occupancy grid, it uses a circular mask to perform a sliding-window match with the target around the supposed center positions that are known by design, until it eventually finds the 4 circular areas with fewer points inside. 
The full target detection algorithm for a LiDAR is explained in Algorithm \ref{pseudocode}.

\subsection{Optimization of sensor poses}\label{subsection:goosp}
The process of multi-sensor calibration begins immediately after the successful recognition of a common target by multiple sensors. Once a target is successfully identified, any sensor that has detected it can be included in the estimation of the relative spatial transformations between all participating sensors. The core of this estimation is an optimization problem that minimizes the reprojection errors of the target's geometric features, specifically the centers of a circular pattern. To solve this problem, the system designates a reference frame, denoted as frame B, which is associated with the first camera to successfully and correctly identify the target. All subsequent measurements from other sensors are then transformed into this unified reference frame. This transformation allows for a direct comparison and a consistent measure of error across the entire sensor suite. The reprojection error, therefore, is calculated and minimized within this common reference frame. This approach ensures that all measurements, regardless of the sensor of origin, are geometrically consistent. The optimization framework is designed to estimate the relative transformation between every sensor and the designated reference frame \textbf{B}.  Every transformation between two sensors can be composed using
\begin{equation}
    T_{L_i}^{C_j} = (T_{C_j}^{B})^{-1} \, T_{L_i}^{B}.
\end{equation}

Assuming a sensor suite made up of $N$ LiDARs and $M$ Cameras, we define as $\rho_{L_i,  C_j}$ the residual component that shapes the error of a relative transformation between the $i^{th}$ LiDAR and the $j_{^th}$ Camera, $\rho_{Ci_,  C_j}$ as the residual between two cameras, and $\rho_{
L_i,  L_j}$ the residual between two LiDARs.
Given $^C\mathbf{p}_i \in \mathbb{R}^3$, the $i^{th}$ center position in the camera frame, and $^L\mathbf{p}_i$ the same measure expressed in LiDAR frame, the optimization problem estimates the relative transformations set $\mathbf{T}_{LC}^B$ that transform every sensor frame to the reference one, that is

\begin{equation*}
    \mathbf{T}_{LC}^B = [T_{C_1}^{B},\dots, T_{C_m}^{B}, T_{L_1}^{B}, \dots, T_{L_n}^{B}]
\end{equation*}

These residuals are minimized to solve the resulting optimization problem:
\begin{equation}
   \mathbf{T}_{LC}^B =  \arg \min_{\mathbf{T^{LC}}} \frac{1}{2} || \rho_{CC} + \rho_{LC} + \rho_{LL}||^2
\end{equation}
with:
\begin{align}
\rho_{CC} &=  \sum_{c_i = 1}^M \sum_{c_j = 1}^M\left(\sum_{k=1}^4 \pi_C(T_B^{C_j} T_{C_i}^B\mathbf{p}_{3D,k}^{C_i}) - \mathbf{p}_{2D,k}^{C_j}\right) \\
\rho_{LC} &=  \sum_{l_i = 1}^N \sum_{c_j = 1}^M\left( \sum_{k=1}^4 \pi_C(T_B^{C_j} T_{L_i}^B\mathbf{p}_{3D,k}^{L_i}) - \mathbf{p}_{2D,k}^{C_j} \right) \\
\rho_{LL} &= \sum_{l_i = 1}^N \sum_{l_j = 1}^N\left( \sum_{k = 1}^4 T_{L_i}^B\, \mathbf{p}^{L_i}_k - T_{L_j}^B\, \mathbf{p}^{L_j}_k \right).
\end{align}
This formulation provides a residual term for every possible pair combination in the sensor suite.

\begin{figure*}
    \centering
    \includegraphics[width=\linewidth]{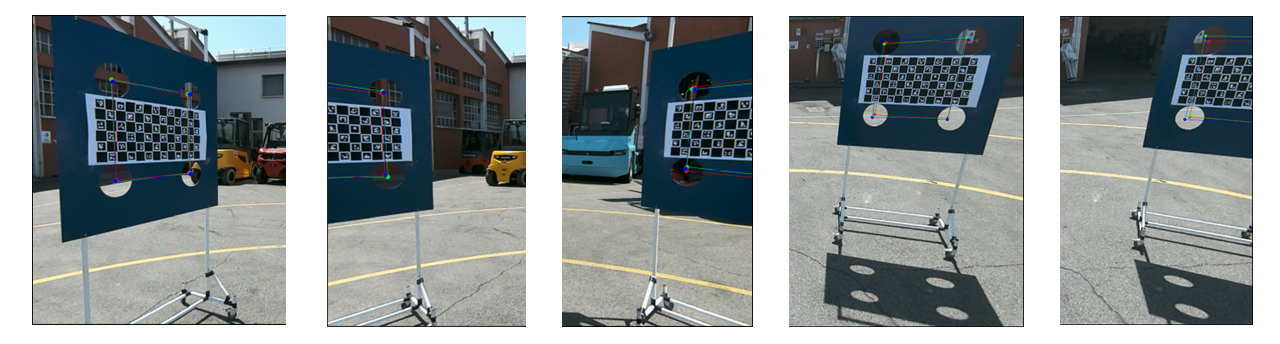}
    \caption{Projection of every center detected by each sensor in the image collected by the reference camera. The red square is related to the first camera that detects the target. The blue and green squares are centers detected by the two LiDARs.}
    \label{fig:different_angles}
\end{figure*}

\begin{figure}
    \centering

    \scriptsize

    \captionof{table}{Compact Calibration Results to Camera 0}
    \begin{tabular}{|l|c|c|}
    
    \hline
    \textbf{Sensor} & \textbf{Translation (x, y, z in meters)} & \textbf{Euler Angles(XYZ in degrees)} \\
    \hline
    LiDAR 0 & (-0.6165, 0.2887, 0.1292) & (110.80, -1.609, -87.738) \\
    LiDAR 1 & (0.6365, 0.2604, 0.1367) & (109.385, 0.6862, 92.213) \\
    Cam 1 & (-0.6600, 0.2836, 0.1508) & (-35.710, -71.299, -48.371) \\
    Cam 2 & (0.6628, 0.2892, 0.1796) & (12.397, 72.015, 7.107) \\
    \hline
    
    \end{tabular}
    \label{tab:excal}

    \vspace{0.2cm}

    \includegraphics[width=0.7\columnwidth]{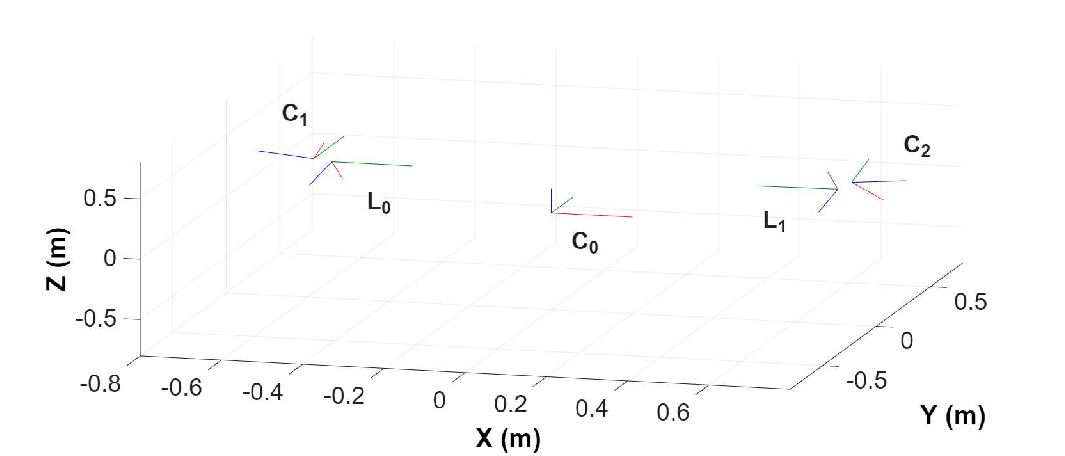} 
    \captionof{figure}{Each sensor disposition in space with respect to the reference one. Frames are rotated and translated using the information in Table I where relative position is expressed in meters and relative orientation in degrees.}
    \label{fig:combined_vertical}
\end{figure}

\begin{figure}[htbp]
    \centering
    \includegraphics[width=0.5\textwidth]{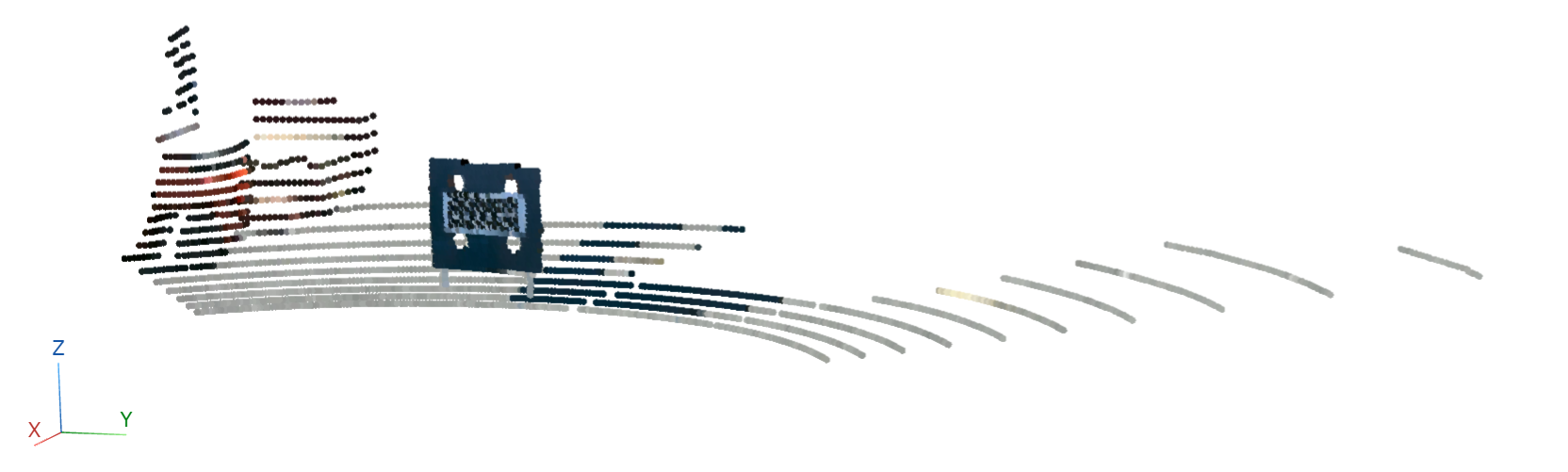}
    \caption{LiDAR pointcloud coloured using camera projection information with the estimated transformation.}
    \label{fig:coloured_pointcloud}
\end{figure}
\newpage

\begin{table} [htbp]
\centering
\caption{Residuals for every sensor pair that detected the target in each sequence. S1, S2, and S3 refer to the three cameras, while S4 and S5 represent the two LiDARs. Each Reprojection error refers to the respective center and are expressed in meters.}
\label{tab:sensor_residuals_improved}
\begin{tabular}{c l S[table-format=1.4]} 
\toprule
\textbf{Sequence} & \textbf{Pair} & {\textbf{Reprojection Errors (m)}} \\
\midrule
1 & S1-S4 & {[0.0698, 0.0653, 0.0817, 0.0967]} \\
& S4-S5 & {[0.0342, 0.0085, 0.0253, 0.0091]} \\
& S5-S1 & {[0.0835, 0.0663, 0.0826, 0.0958]} \\
\midrule
2 & S1-S4 & {[0.0670, 0.0682, 0.0849, 0.0879]} \\
& S4-S5 & {[0.0360, 0.0299, 0.0150, 0.0128]} \\
& S5-S1 & {[0.0776, 0.0616, 0.0737, 0.0813]} \\
\midrule
3 & S1-S4 & {[0.0838, 0.0843, 0.0833, 0.0888]} \\
& S4-S5 & {[0.0399, 0.0207, 0.0226, 0.0167]} \\
& S5-S1 & {[0.0806, 0.0716, 0.0640, 0.0798]} \\
\midrule
4 & S1-S3 & {[0.0175, 0.0138, 0.0352, 0.0383]} \\
& S1-S4 & {[0.0863, 0.0758, 0.0870, 0.0767]} \\
& S1-S5 & {[0.0599, 0.0534, 0.0563, 0.0677]} \\
& S3-S4 & {[0.0988, 0.0869, 0.1181, 0.1137]} \\
& S3-S5 & {[0.0728, 0.0648, 0.0896, 0.0976]} \\
& S4-S5 & {[0.0303, 0.0236, 0.0327, 0.0440]} \\
\midrule
5 & S3-S4 & {[0.0960, 0.1067, 0.0837, 0.0744]} \\
& S3-S5 & {[0.0733, 0.0820, 0.0549, 0.0573]} \\
& S4-S5 & {[0.0230, 0.0412, 0.0341, 0.0331]} \\
\midrule
6 & S3-S4 & {[0.0942, 0.0967, 0.0940, 0.0916]} \\
& S3-S5 & {[0.0666, 0.0702, 0.0760, 0.0633]} \\
& S4-S5 & {[0.0385, 0.0304, 0.0327, 0.0357]} \\
\midrule
7 & S3-S5 & {[0.0914, 0.0914, 0.0809, 0.0802]} \\
\midrule
8 & S3-S5 & {[0.0914, 0.0987, 0.0893, 0.0801]} \\
\midrule
9 & S3-S4 & {[0.0963, 0.1066, 0.1153, 0.1149]} \\
& S3-S5 & {[0.0790, 0.0848, 0.0843, 0.0825]} \\
& S4-S5 & {[0.0382, 0.0313, 0.0333, 0.0339]} \\
\midrule
10 & S3-S4 & {[0.0658, 0.0928, 0.1198, 0.0837]} \\
& S3-S5 & {[0.0495, 0.0761, 0.0816, 0.0605]} \\
& S4-S5 & {[0.0210, 0.0253, 0.0391, 0.0236]} \\

\bottomrule
\label{table:residuals}
\end{tabular}

\end{table}
\section{Experiments} \label{section:experiments}
The proposed approach has been tested on a vehicle equipped with 3 STURDeCAM25 Cameras and 2 Hesai LiDARs. Twenty sequences were systematically collected from the external perimeter of a warehouse facility, with the target positioned in various locations to ensure detection by all sensors, as shown in Figure \ref{fig:different_angles}. The changes in the target's pose ensure that it is visible to each sensor at least once, which is sufficient for the optimization framework to determine each sensor's position relative to the others. Due to the vehicle's dimensions and the complex sensor positioning, we were not able to measure the ground truth configuration. The calibration software, developed in C++ within the ROS framework, leverages the Ceres library\cite{garwal_Ceres_Solver_2022} to perform optimization. The solver employs dense Cholesky factorization and is configured to run for a maximum of 5000 iterations. No particular loss function or residual weighting policies have been used to solve the problem. First, we carry out a consistency check on the obtained transformations: we compose the transformations to make sure that the resulting transformation coincides with the initial reference frame, that is, starting from the frame related to the first camera:
\begin{equation*}
    {C_0}\xrightarrow{T_{C_0}^{C_1}}{C_1}\xrightarrow{T_{C_1}^{C_2}}{C_2}\xrightarrow{T_{C_2}^{CL_0}}{L_0}\xrightarrow{T_{L_0}^{L_1}}{L_1}\xrightarrow{T_{L_1}^{C_0}}{C_0} 
\end{equation*}
The output of this chain of transformation is an Identity matrix, which confirms the consistency of the estimated poses. We then project the centers detected by each sensor in the reference frame $B$ of the first camera that detected it, as explained in Section \ref{subsection:goosp}. This allows for measuring the accuracy of the estimated transformation by looking at the Euclidean distance of every center with respect to the ones detected by the reference camera.

\section{Results}
As explained in Section \ref{section:experiments}, to evaluate the final accuracy of the described system, we projected the estimated circle centers in a reference frame (Figure \ref{fig:lidarcamera_centers_projection}) as described in Section \ref{section:ourapproach}, and we measured the reprojection errors between them as in Table \ref{table:residuals}. Most SLAM, calibration, and tracking software relying on bundle adjustment or pose graph optimization will employ reprojection error as a key loss or residual term\cite{campos2021orb, frosi2022mcs}. Moreover, in order to have a visual feedback from the LiDARs estimated transformation, in Figure \ref{fig:coloured_pointcloud} we coloured the pointclouds with the image colours using the estimated trajectory. To assess the performance under challenging conditions, the target was rotated multiple times throughout the sequences, thereby introducing a variety of angles, as shown in Figure \ref{fig:different_angles}. This procedure enables a rigorous evaluation of the precision of the estimated transformations. It is important to note that, due to the specific design of the target board and the lack of overlapping fields of view among the cameras, a direct comparison with existing state-of-the-art methods is not straightforward\cite{yan2022joint, lee2020unified}. The estimated transformations of each sensor with respect to the reference frame (i.e. Camera 0) are listed in Table \ref{tab:excal} and Figure \ref{fig:combined_vertical}.

\section{Conclusions}
In this paper, we presented a multi-LiDAR multi-camera extrinsic calibration system that leverages a custom target board. Through an optimization problem, the system computes the transformation required to align each sensor measurement. The proposed method systematically exploits both geometric features from LiDAR data and visual cues from camera images, ensuring robust and accurate estimation of inter-sensor transformations. By integrating all available sensor observations into a unified optimization framework, the approach achieves consistent calibration across the entire sensor suite, regardless of modality or number. Experimental validation demonstrates that the system produces precise and repeatable results.

\begin{credits}

\subsubsection{\discintname}
The authors have no competing interests to declare that are
relevant to the content of this article.
\end{credits}
%
%
%
%
\bibliographystyle{splncs04}
\bibliography{bibliography}

@ARTICLE{shahbeigi2024,
  author={Shahbeigi, Sepeedeh and Robinson, Jonathan and Donzella, Valentina},
  journal={IEEE Access}, 
  title={A Novel Score-Based LiDAR Point Cloud Degradation Analysis Method}, 
  year={2024},
  volume={12},
  number={},
  pages={22671-22686},
  doi={10.1109/ACCESS.2024.3359300}}

@article{du2025,
title = {Robust Point Cloud Registration Based on Semantic Iterative Closest Point Algorithm},
journal = {Fundamental Research},
year = {2025},
issn = {2667-3258},
doi = {https://doi.org/10.1016/j.fmre.2024.04.025},
author = {Shaoyi Du and Tiancheng Shao and Canhui Tang and Wei Zeng and Zhiqiang Tian}
}

@article{ma2021crlf,
  title={CRLF: Automatic calibration and refinement based on line feature for LiDAR and camera in road scenes},
  author={Ma, Tao and Liu, Zhizheng and Yan, Guohang and Li, Yikang},
  journal={arXiv preprint arXiv:2103.04558},
  year={2021}
}

@ARTICLE{lin2025,
  author={Lin, Ping-Tzu and Huang, Ying-Shiuan and Lin, Wen-Chieh and Wang, Chieh-Chih and Lin, Huei-Yung},
  journal={IEEE Open Journal of Intelligent Transportation Systems}, 
  title={Online LiDAR-Camera Extrinsic Calibration Using Selected Semantic Features}, 
  year={2025},
  volume={6},
  number={},
  pages={456-464},
  doi={10.1109/OJITS.2025.3555574}}

@inproceedings{kirillov2023segment,
  title={Segment anything},
  author={Kirillov, Alexander and Mintun, Eric and Ravi, Nikhila and Mao, Hanzi and Rolland, Chloe and Gustafson, Laura and Xiao, Tete and Whitehead, Spencer and Berg, Alexander C and Lo, Wan-Yen and others},
  booktitle={Proceedings of the IEEE/CVF international conference on computer vision},
  pages={4015--4026},
  year={2023}
}

@inproceedings{gicp2009,
  title={Generalized-icp.},
  author={Segal, Aleksandr and Haehnel, Dirk and Thrun, Sebastian},
  booktitle={Robotics: science and systems},
  volume={2},
  number={4},
  pages={435},
  year={2009},
  organization={Seattle, WA}
}

@ARTICLE{yin2023,
  author={Yin, Jun and Yan, Fei and Liu, Yisha and Zhuang, Yan},
  journal={IEEE Sensors Journal}, 
  title={Automatic and Targetless LiDAR–Camera Extrinsic Calibration Using Edge Alignment}, 
  year={2023},
  volume={23},
  number={17},
  pages={19871-19880},
  doi={10.1109/JSEN.2023.3297522}}

@article{kang2020automatic,
  title={Automatic targetless camera-lidar calibration by aligning edge with gaussian mixture model},
  author={Kang, Jaehyeon and Doh, Nakju L},
  journal={Journal of Field Robotics},
  volume={37},
  number={1},
  pages={158--179},
  year={2020},
  publisher={Wiley Online Library}
}

@inproceedings{zhou2018automatic,
  title={Automatic extrinsic calibration of a camera and a 3d lidar using line and plane correspondences},
  author={Zhou, Lipu and Li, Zimo and Kaess, Michael},
  booktitle={2018 IEEE/RSJ International Conference on Intelligent Robots and Systems (IROS)},
  pages={5562--5569},
  year={2018},
  organization={IEEE}
}

@article{an2024survey,
  title={Survey of extrinsic calibration on lidar-camera system for intelligent vehicle: Challenges, approaches, and trends},
  author={An, Pei and Ding, Junfeng and Quan, Siwen and Yang, Jiaqi and Yang, You and Liu, Qiong and Ma, Jie},
  journal={IEEE Transactions on Intelligent Transportation Systems},
  year={2024},
  publisher={IEEE}
}

@Article{park2014,
AUTHOR = {Park, Yoonsu and Yun, Seokmin and Won, Chee Sun and Cho, Kyungeun and Um, Kyhyun and Sim, Sungdae},
TITLE = {Calibration between Color Camera and 3D LIDAR Instruments with a Polygonal Planar Board},
JOURNAL = {Sensors},
VOLUME = {14},
YEAR = {2014},
NUMBER = {3},
PAGES = {5333--5353},
URL = {https://www.mdpi.com/1424-8220/14/3/5333},
PubMedID = {24643005},
ISSN = {1424-8220},
DOI = {10.3390/s140305333}
}

@inproceedings{yaopeng2021review,
  title={Review of a 3D LiDAR combined with single vision calibration},
  author={Yaopeng, Liu and Xiaojun, Guo and Shaojing, Su and Bei, Sun},
  booktitle={2021 IEEE International Conference on Data Science and Computer Application (ICDSCA)},
  pages={397--404},
  year={2021},
  organization={IEEE}
}

@article{beltran2022automatic,
  title={Automatic extrinsic calibration method for lidar and camera sensor setups},
  author={Beltr{\'a}n, Jorge and Guindel, Carlos and De La Escalera, Arturo and Garc{\'\i}a, Fernando},
  journal={IEEE Transactions on Intelligent Transportation Systems},
  volume={23},
  number={10},
  pages={17677--17689},
  year={2022},
  publisher={IEEE}
}

@inproceedings{fischler1981random,
  title={Random sample consensus: a paradigm for model fitting with applications to image analysis and automated cartography},
  author={Fischler, Martin A and Bolles, Robert C},
  booktitle={Communications of the ACM},
  volume={24},
  number={6},
  pages={381--395},
  year={1981},
  publisher={ACM}
}

@article{hsu2020review,
  title={A review and perspective on optical phased array for automotive LiDAR},
  author={Hsu, Ching-Pai and Li, Boda and Solano-Rivas, Braulio and Gohil, Amar R and Chan, Pak Hung and Moore, Andrew D and Donzella, Valentina},
  journal={IEEE Journal of Selected Topics in Quantum Electronics},
  volume={27},
  number={1},
  pages={1--16},
  year={2020},
  publisher={IEEE}
}

@article{li2020lidar,
  title={Lidar for autonomous driving: The principles, challenges, and trends for automotive lidar and perception systems},
  author={Li, You and Ibanez-Guzman, Javier},
  journal={IEEE Signal Processing Magazine},
  volume={37},
  number={4},
  pages={50--61},
  year={2020},
  publisher={IEEE}
}

@article{campos2021orb,
  title={Orb-slam3: An accurate open-source library for visual, visual--inertial, and multimap slam},
  author={Campos, Carlos and Elvira, Richard and Rodr{\'\i}guez, Juan J G{\'o}mez and Montiel, Jos{\'e} MM and Tard{\'o}s, Juan D},
  journal={IEEE transactions on robotics},
  volume={37},
  number={6},
  pages={1874--1890},
  year={2021},
  publisher={IEEE}
}

@inproceedings{frosi2022mcs,
  title={MCS-SLAM: Multi-Cues Multi-Sensors Fusion SLAM},
  author={Frosi, Matteo and Matteucci, Matteo},
  booktitle={2022 IEEE Intelligent Vehicles Symposium (IV)},
  pages={1423--1429},
  year={2022},
  organization={IEEE}
}

@article{yan2022joint,
  title={Joint camera intrinsic and lidar-camera extrinsic calibration},
  author={Yan, Guohang and He, Feiyu and Shi, Chunlei and Cai, Xinyu and Li, Yikang},
  journal={arXiv preprint arXiv:2202.13708},
  year={2022}
}

@inproceedings{lee2020unified,
  title={Unified calibration for multi-camera multi-lidar systems using a single checkerboard},
  author={Lee, Wonmyung and Won, Changhee and Lim, Jongwoo},
  booktitle={2020 IEEE/RSJ International Conference on Intelligent Robots and Systems (IROS)},
  pages={9033--9039},
  year={2020},
  organization={IEEE}
}

@article{an2018charuco,
  title={Charuco board-based omnidirectional camera calibration method},
  author={An, Gwon Hwan and Lee, Siyeong and Seo, Min-Woo and Yun, Kugjin and Cheong, Won-Sik and Kang, Suk-Ju},
  journal={Electronics},
  volume={7},
  number={12},
  pages={421},
  year={2018},
  publisher={MDPI}
}

@software{garwal_Ceres_Solver_2022,
  author = {Agarwal, Sameer and Mierle, Keir and The Ceres Solver Team},
  title = {{Ceres Solver}},
  license = {Apache-2.0},
  url = {https://github.com/ceres-solver/ceres-solver},
  version = {2.2},
  year = {2023},
  month = {10}
}
\end{document}